\pdfoutput=1

\documentclass[11pt]{article}

\usepackage[]{acl}

\usepackage{times}
\usepackage{latexsym}

\usepackage[T1]{fontenc}

\usepackage[utf8]{inputenc}

\usepackage{microtype}

\usepackage{booktabs, colortbl} 
\usepackage{multirow,makecell} 
\usepackage{multicol}
\usepackage{amsmath}
\usepackage{amssymb}
\usepackage{caption}
\usepackage{subcaption}
\usepackage{titlesec}
\usepackage{pgfplots}
\usepackage[normalem]{ulem} 
\usepackage{enumitem}
\usepackage{graphicx} 
\usepackage{hyperref}

\usepackage{inconsolata}

\title{How Transferable are Attribute Controllers on Pretrained Multilingual Translation Models?}

\author{Danni Liu \and Jan Niehues \\
        Karlsruhe Institute of Technology, Germany \\
        \texttt{\{danni.liu, jan.niehues\}@kit.edu}} 

\begin{document}
\maketitle
\begin{abstract}
Customizing machine translation models to comply with desired attributes (e.g., formality or grammatical gender) is a well-studied topic.
However, 
most current approaches rely on (semi-)supervised data with attribute annotations.
This data scarcity bottlenecks democratizing such customization possibilities to a wider range of languages, particularly lower-resource ones.
This gap is out of sync with recent progress in pretrained massively multilingual translation models.
In response, we transfer the attribute controlling capabilities to languages without attribute-annotated data with an NLLB-200 model as a foundation.
Inspired by techniques from controllable generation, 
we employ a gradient-based inference-time controller to steer the pretrained model. 
The controller transfers well to zero-shot conditions, 
as it operates on pretrained multilingual representations and is attribute- rather than language-specific.
With a comprehensive comparison to finetuning-based control,
we demonstrate that,
despite finetuning's clear dominance in supervised settings, the gap to inference-time control closes when moving to zero-shot conditions, 
especially with new and distant target languages.
The latter also shows stronger domain robustness. 
We further show that our inference-time control complements finetuning.
A human evaluation on a real low-resource language, Bengali, confirms our findings.
Our code is \href{https://github.com/dannigt/attribute-controller-transfer}{here}.
\end{abstract}

\section{Introduction}
\begin{figure}%
\centering
\includegraphics[trim={0cm 0cm 0cm 0cm},clip,width=\linewidth]{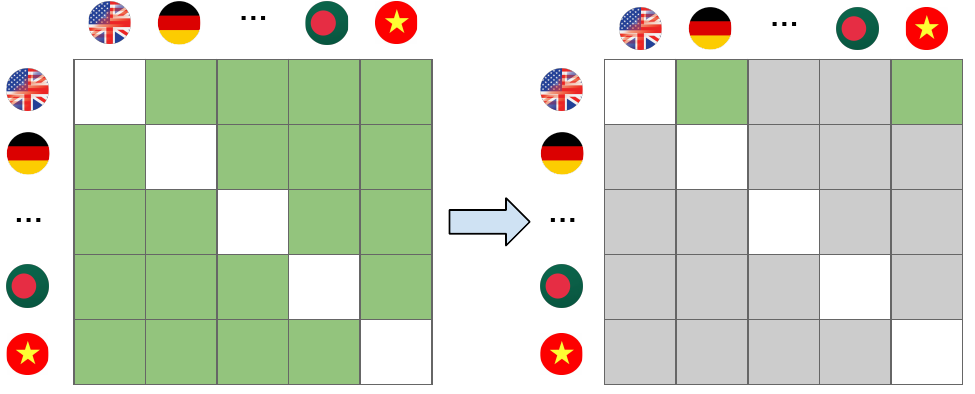}
\caption{\label{fig:data_condition} 
The number of translation directions with attribute-annotated data (\textbf{right}) is far less than that of what massively pretrained models serve (\textbf{left}).
}
\vspace{-8pt}
\end{figure}

Pretrained multilingual translation models with massive coverage~\cite{zhang-etal-2020-improving,liu-etal-2020-multilingual-denoising, m2m100, xue-etal-2021-mt5, nllb} have become of the backbone of many translation systems. 
While their off-the-shelf translation quality has been constantly improving~\cite{m2m100, deltalm, nllb},
the flexibility of customization towards desired attributes, such as formality or grammatical gender, is another important metric.
Adapting generic systems for attribute-controlled translation relies on training data with attribute information.
Creating such annotated data often requires language-specific knowledge and manual curation. 
This makes data acquisition challenging even for single languages.
When scaling to the numerous directions served by massively multilingual models, it quickly becomes impractical, as shown in \autoref{fig:data_condition}.
While prior works~\cite{michel-neubig-2018-extreme,saunders-etal-2020-neural,nadejde-etal-2022-cocoa} showed promising results of finetuning on limited attribute-annotated data,
to allow other languages without supervised data to similarly benefit from the customization possibilities,
the \textit{transferability} of the attribute controllers remains to be studied.

\begin{figure*}%
\centering
\includegraphics[trim={0cm 0.25cm 0.1cm 0.09cm},clip,width=\linewidth]{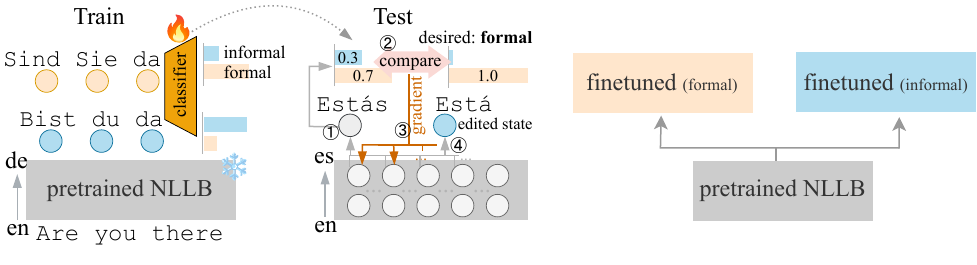}
\caption{\label{fig:intro} 
Left: \textbf{Inference}-time control by gradient-based classifier guidance: training classifiers for attributes on decoder activations, 
and using its predictions to edit inference-time model \textit{activations} towards desired attributes.
Right: Standard \textbf{training}-time control by finetuning on attribute-specific data.
}
\vspace{-6pt}
\end{figure*}

A straightforward way to achieve attribute control is finetuning on attribute-specific data.
Recent works~\cite{rippeth-etal-2022-controlling,wu-etal-2023-improving} have shown that finetuning with just hundreds of attribute-specific sentences is sufficient.
However, small finetuning data also brings the risk of overfitting and catastrophic forgetting~\cite{DBLP:journals/corr/FreitagA16,thompson-etal-2019-overcoming}.
It is especially relevant when generalizing to new languages,
where finetuning on some languages may erase the knowledge of others from pretraining~\cite{garcia-etal-2021-towards,cooper-stickland-etal-2021-multilingual,liu-niehues-2022-learning}.
While these issues may be mitigated by partial finetuning \cite{DBLP:conf/icml/HoulsbyGJMLGAG19,bapna-firat-2019-simple}, 
domain mismatch between the finetuning data and the test domain can still degrade translation quality.
We will validate these concerns in various zero-shot conditions with different language relatedness and domains.

On the other end of the spectrum, 
inference-time customization is another paradigm of attribute control.
In this case, the pretrained model is fully unchanged in the training stage.
At inference time, the generation process is steered towards desired attributes by 
e.g. re-weighting entries in the output distribution~\cite{saboo-baumann-2019-integration,yang-klein-2021-fudge,landsman-etal-2022-beamr} 
or editing model activations~\cite{DBLP:conf/iclr/DathathriMLHFMY20}.
To enable cross-lingual transfer, the controller must be trained on features that are shared across languages.
This precludes methods that operate on the \textit{surface} vocabulary level. 
In this work, 
we will extend an activation-based approach ~\cite{DBLP:conf/iclr/DathathriMLHFMY20} originally for decoder-only models to cross-lingual transfer on pretrained translation models.

\noindent
\textbf{Task Formalization}
We focus on the following task:
Given a pretrained many-to-many multilingual translation model covering $N$ languages and $N(N-1)$ translation directions, along with parallel data on $k$ ($k \ll N(N-1)$) translation directions where the target translation corresponds to specific attributes (e.g., formality level),
we aim to customize the pretrained model to translate with desired attributes for as many directions as possible.
We refer to the subsequent model as an \textit{attribute controller}.
Specifically, 
after learning on the $k$ sets of parallel data with attribute annotation,
to what extent can we transfer the attribute controller to the remaining $N(N-1) - k$ translation directions?
 
\section{Background and Related Work}
\noindent
\textbf{Attribute-Controlled Translation}
Previous works investigated controlling various attributes of machine translation outputs, for instance politeness \cite{sennrich-etal-2016-controlling,niu-etal-2018-multi,feely-etal-2019-controlling}, gender \cite{vanmassenhove-etal-2018-getting,saunders-etal-2020-neural}, length \cite{takase-okazaki-2019-positional,lakew-etal-2019-controlling,marchisio-etal-2019-controlling,niehues-2020-machine}, or style in general \cite{michel-neubig-2018-extreme,schioppa-etal-2021-controlling,vincent-etal-2023-mtcue,wang-etal-2023-controlling}.
As existing works mainly focus on supervised conditions with at least some supervised data, how these approaches generalize to new languages remains unclear.
In face of data scarcity, one approach is to use synthetic data by pseudo-labeling \cite{rippeth-etal-2022-controlling,lee-etal-2023-improving-formality}.
In our work, 
by building upon massively multilingual translation models, 
we \textit{do not assume} the scalability of creating synthetic data for all languages served by the backend model,
\textit{nor do we assume} a classifier that can a priori distinguish attribute classes for zero-shot languages. 
%
 
\noindent
\textbf{Multilinguality for Controllable Generation}
Our work is also related to controllable text generation in general.
Despite steady progress in this field \cite{DBLP:journals/corr/abs-1909-05858,krause-etal-2021-gedi-generative,yang-klein-2021-fudge,liu-etal-2021-dexperts}, 
how the controller generalizes across languages is likewise less explored. 
With the recent surge of large language models (LLMs), attribute-controlled translation has also been addressed by prompting multilingual language models in a few-shot manner \cite{sarti-etal-2023-ramp,DBLP:conf/icml/GarciaBCFKJF23}.
Notably, \citet{sarti-etal-2023-ramp} reported promising few- and zero-shot attribute control results using multilingual LLMs.
In this work, 
we take a different perspective by using a pretrained dedicated encoder-decoder translation model as backend, 
and transferring the attribute control capabilities with lightweight add-ons.
As currently open LLMs still lag behind dedicated translation models \cite{DBLP:journals/corr/abs-2304-04675,sarti-etal-2023-ramp} especially on low-resource languages \cite{DBLP:journals/corr/abs-2309-07423},
we believe improving the attribute control capabilities of massively multilingual conventional models is still highly relevant.

\noindent
\textbf{Multilingual Domain Adaptation} Attribute control can be viewed as a light domain adaptation task.
 Prior works \cite{cooper-stickland-etal-2021-multilingual,vu-etal-2022-domains} adapting pretrained multilingual models have reported catastrophic forgetting of languages absent from the finetuning stage.
 Our results on finetuning for zero-shot attribute control (\S\ref{subsec:new_target}) shows a different picture.
 One potential reason is that, compared to adapting to fully new domains such as medical or law texts, 
 the attribute control task can be learned with less data.
 This in turn requires less intense finetuning and is therefore less vulnerable to forgetting.

\section{Transferring Attribute Controllers for Multilingual Translation}
To generalize to new translation directions,
an ideal controller should be \textit{attribute}- rather than \textit{language}-specific.
That is, 
its representation for different attribute labels varies little with specific languages.

\paragraph{Inference-Time Control by Classifier Guidance:} \label{subsec:cg_approach}
Our first approach builds upon the observation that the activations of pretrained multilingual models capture commonalities of different languages~\cite{pires-etal-2019-multilingual,liu-etal-2020-multilingual-denoising}.
An attribute classifier trained on these activations can then potentially transfer across languages, 
which we use at inference time to steer the generation for languages without attribute-annotated data.
The control takes effect on inference-time model \textit{activations} instead of \textit{parameters}, as shown in \autoref{fig:intro}.
Specifically, 
we first train an attribute classifier while \textit{freezing} the pretrained model, 
and then edit the model activations towards the wanted attribute based on the predicted label at inference time.
This idea has shown success in controllable image synthesis~\cite{classifier_guidance} and text generation~\cite{DBLP:conf/nips/LiTGLH22}.
To the best of our knowledge, no prior work has explored it for cross-lingual transfer.

Specifically, 
we extend the approach by \citet{DBLP:conf/iclr/DathathriMLHFMY20} to encoder-decoder models.
For machine translation, 
Given a frozen pretrained model, 
we run forward passes with attribute-annotated\footnote{Only the target side needs attribute labels.} parallel data $(\mathbf{X}, \mathbf{Y})^c$ 
for $c \in [C]$, where $\mathbf{X}$ and $\mathbf{Y}$ are the source and target sentences with individual sentence pairs $(\mathbf{x}, \mathbf{y})^{i} \in (\mathbf{X}, \mathbf{Y})$, and $C$ is the number of attribute labels.

{
While freezing the translation model’s parameters,
we train a classifier that maximizes $P(c \mid \mathbf{h})$,
where $c$ is the ground-truth attribute label and $\mathbf{h}$ is the last decoder layer's hidden states after forced-decoding parallel data $(\mathbf{x}, \mathbf{y})$:
\begin{equation}
    \mathbf{h} =\text{decoder}(\mathbf{y}, \text{encoder}(\mathbf{x})).
\end{equation}
Like with a standard model, the output distribution is then 
$\text{softmax}(\mathbf{W} \mathbf{h})$,
where $\mathbf{W}$ maps the hidden states $\mathbf{h}$ to the vocabulary distribution.

At inference time step $t$, the hidden state is:
\begin{equation}
\mathbf{h}_{t} =\text{decoder}(y_{t-1}, \mathbf{A}_{t-1}), 
\end{equation}
where $y_{t-1}$ is the token from the previous step, 
and $\mathbf{A}_{t-1}$ is the model activations.
$\mathbf{A}_{t-1}$ contains activation key-value pairs\footnote{Note these are not the key/value projection weights of the Transformer, but the activations after applying the projections.} 
from the decoder self-attention and cross-attention for steps $1$ to $t-1$, 
and is cached in most Transformer decoding implementations~\cite{ott-etal-2019-fairseq,wolf-etal-2020-transformers}.

Based on all available decoder states till $t-1$, 
we predict an attribute label:
    $\mathrm{argmax}_c P(c \mid \mathbf{h}_{1,...,t-1})$.
Following \citet{DBLP:conf/iclr/DathathriMLHFMY20}, we meanpool the states from timestep $1$ to $t-1$ for the prediction.
It also empirically showed better performance than 1) using a token-level classifier without pooling and 2) operating on the cumulative sum of hidden states from all time steps so far.\footnote{In initial experiments training an English-German formality classifier, the accuracy on the dev set was 86.7\% (meanpool), 66.1\% (token-level) and 73.0\% (cumulative sum).}

As $\mathbf{h}_{1,...,t-1}$ is only determined by $\mathbf{A}_{t-1}$,
we can rewrite $P(c \mid \mathbf{h}_{1,...,t-1})$ as $P(c \mid \mathbf{A}_{t-1})$.
Comparing the prediction to the desired attribute $c^*$, 
we can derive gradients measuring how much the current activations satisfy the desired $c^*$.
The gradients, $\nabla_{\mathbf{A}_{t-1}}P(c^* \mid \mathbf{A}_{t-1})$, 
are then back-propagated for several iterations with given step sizes,
resulting in updated activations $\Tilde{\mathbf{A}}_{t-1}$, 
which further leads to modified decoder hidden state:
\begin{equation}
\Tilde{\mathbf{h}}_{t} =\text{decoder}(y_{t-1}, \Tilde{\mathbf{A}}_{t-1}).
\end{equation}
A new output token $y_t$ (that more likely satisfies the control) is generated from $\Tilde{\mathbf{h}}_{t}$ by $\text{softmax}(\mathbf{W} \Tilde{\mathbf{h}}_{t})$.
}

\paragraph{Finetuning-Based Control:}
A more common way to realize control is finetuning the pretrained model on attribute-specific parallel data, as done in domain adaptation \cite{DBLP:journals/corr/FreitagA16}.
To transfer to directions without annotated data, 
the adaptation step must mostly learn the desired \textit{attributes} rather than the specific \textit{languages} in finetuning, 
so as not to forget the languages without annotated data.
On our tasks, naive finetuning already works effectively:
We finetune the full model on each attribute, 
resulting in one specialized model per attribute as shown in \autoref{fig:intro}.\footnote{We tried prepending attribute tags to the source sentences~\cite{chu-etal-2017-empirical,DBLP:conf/ranlp/KobusCS17},
but this was not enough to make the pretrained model to be attribute-aware.
A potential reason is that the pretrained model tends ignore the source tags as noise, and that the low amount of finetuning data cannot re-establish the importance of the tags.}
Partial finetuning e.g. with adapters \cite{bapna-firat-2019-simple,philip-etal-2020-monolingual} is a more parameter-efficient approach.
We do not explore partial finetuning in this work, 
as it does not fully align with our focus on the transferability of attribute controllers.

\begin{table}[t]
\small
\centering
\setlength\tabcolsep{0pt} 
\begin{tabular}{lllrr}
\toprule
\multicolumn{2}{l}{\textbf{Task}} & 
\textbf{Directions} & 
\multicolumn{2}{r}{\textbf{\# Sent. per lang. per att.}} \\
\midrule
\multicolumn{4}{l}{\textbf{Formality control} (formal/informal)} \\
&
train & 
\multicolumn{2}{l}{en$\rightarrow$\{de, es, fr, hi, it\}
} 
&
400
\\
& test (supervised)
& \multicolumn{2}{l}{en$\rightarrow$\{de, es, fr, hi, it\}
}
& 600
\\
& test (new tgt)
& \multicolumn{2}{l}{en$\rightarrow$\{pt, ru, ko\}
}
& 600
\\
&test (new src)
& \multicolumn{2}{l}{\{de, fr, hi, it\}$\rightarrow$es
}
&366-572
\\
\multicolumn{4}{l}{\textbf{Grammatical gender control} (feminine/masculine)} \\
& train 
&  \multicolumn{2}{l}{en$\rightarrow$es}
& 194
\\
&
test (supervised)
& 
 \multicolumn{2}{l}{en$\rightarrow$es}
& 552-556 
\\
&
test (new tgt)
&  \multicolumn{2}{l}{en$\rightarrow$\{it, fr\}}
& 515-546
\\
&
test (new src$+$tgt)
& \multicolumn{2}{l}{\{es, fr\}$\rightarrow$it, \{es, it\}$\rightarrow$fr}
& 271-365
\\
\bottomrule
\end{tabular}
\caption{\label{tab:data_overview} 
Data overview. Codes: German (de), Spanish (es), French (fr), Hindi (hi), Italian (it), Korean (ko), Portuguese (pt), Russian (ru), source (src), target (tgt).
}
\end{table}


\section{Experimental Setup} \label{sec:setup}
We experiment on two attribute control tasks: formality and grammatical gender control.
As outlined in \autoref{tab:data_overview}, the training data has English on the source side.
For the target languages, there is one set of translations per attribute.
The low data volume not only reflects the practical challenge of data acquisition, but is also an established condition in existing benchmarks \cite{nadejde-etal-2022-cocoa}.

\subsection{Formality Control (In-Domain)} \label{subsec:data_formality}

The training data come from CoCoA-MT~\cite{nadejde-etal-2022-cocoa}\footnote{We excluded Japanese, 
where our pretrained model has very low translation accuracy on formality-annotated words ($<$40\%, whereas all 5 other languages score $>$60\%).},
where the test domain overlaps with training.
For zero-shot conditions, we transfer controllers trained on different language pairs to new translation directions.
Specifically, we investigate the following two cases:

\paragraph{Transfer to New Target Languages}
We train the attribute controllers on one or multiple target languages to assess the impact of multilinguality on transfer. 
We compare the following settings:
\begin{itemize}[nolistsep,leftmargin=*]
    \item \textbf{Single-direction}: We use \texttt{en}$\rightarrow$\texttt{es} and \texttt{de}
    as representative Romance and Germanic languages;
    \item \textbf{Multilingual}: We train on all languages in the training data: \texttt{en}$\rightarrow$\{\texttt{de}, \texttt{es}, \texttt{fr}, \texttt{hi}, \texttt{it}\}.
\end{itemize}

For the new target languages, we choose three directions from the IWSLT 2023 formality control shared task\footnote{\url{https://github.com/amazon-science/contrastive-controlled-mt/tree/main/IWSLT2023}}~\cite{iwslt2023_findings}: 
\texttt{en}$\rightarrow$\texttt{pt} (\textbf{close}),
\texttt{en}$\rightarrow$\texttt{ru} (\textbf{related}),
and \texttt{en}$\rightarrow$\texttt{ko} (\textbf{distant})
for their different degrees of relatedness to the languages in training.
Among them, 
\texttt{en}$\rightarrow$\texttt{ko} has 400 sentences of supervised data.
We use it to establish the oracle performance in the presence of supervised data.

\paragraph{Transfer to New Source Languages}
We re-align the CoCoA-MT test set using English as pivot, 
creating a new test set with non-English source and target sentences.\footnote{The original test sets only have English input. 
As the English sentences mostly overlap, 
we create new pairs of two non-English languages by matching their English translations.}
Unlike translating from English, 
here the source sentences also contain formality information.
This allows testing if the model can: 1) \textit{preserve} the source formality level; 2) \textit{change} the source formality level when steered so.

\subsection{Gender Control (Out-of-Domain)} \label{subsec:gender_data}
For the formality control setup above, 
the data for training the attribute controller come from the same domain as the test set.
To evaluate domain generalization, 
for grammatical gender control, 
we train the controller on texts with very different styles from the test data.
For \textit{training} the attribute controller, 
we use the \texttt{en}-\texttt{es} set from \citet{saunders-etal-2020-neural}\footnote{\url{https://github.com/DCSaunders/tagged-gender-coref\#adaptation-sets}} with artificial sentences of very simple grammatical structure up to 7 words. 
In contrast, 
for the \textit{test} set
we use MuST-SHE~\cite{bentivogli-etal-2020-gender}, 
which consists of TED talks with much longer sentences and more versatile styles.
More dataset details are Appendix \ref{appendix:data_domain_mismatch}.
Besides transfer to new target languages like previously (\S \ref{subsec:data_formality}), we also explore the following setting:

\paragraph{Transfer to New Source \& Target Languages}
The MuST-SHE test set comes in \texttt{en}-\{\texttt{es}, \texttt{fr}, \texttt{it}\}.
Like previously, we re-align them using English as pivot, creating non-English source and target sentences.
In this case, both the source and target sentences have the same gender. 
As the attribute training data is in \texttt{en}$\rightarrow$\texttt{es}, 
we evaluate \{\texttt{es}, \texttt{fr}\}$\rightarrow$\texttt{it} and \{\texttt{es}, \texttt{it}\}$\rightarrow$\texttt{fr} for the transfer to new translation directions where both the source and target languages differ from training.

\subsection{Models and Evaluation}
\noindent
\textbf{Models}
We use two types of backend models. 
For the main experiments, 
we use the pretrained NLLB-200 distilled 600M model~\cite{nllb}, which covers 200 languages for many-to-many translation.
We also train a Transformer-base~\cite{transformer} from scratch to verify if observed phenomena are specific to models with massive multilingual pretraining.
The Transformer-base model covers all languages in our experiments and is trained on OPUS-100~\cite{zhang-etal-2020-improving}.
Details of these data are in Appendix \ref{appendix:opus_data}.
Training and inference details are in \autoref{appendix:training_inference}.

\noindent
\textbf{Control Evaluation}
For \textit{formality} control, we report matched accuracy (M-Acc; \%) following \citet{nadejde-etal-2022-cocoa}.
For \textit{gender} control, we use the official evaluation script~\cite{bentivogli-etal-2020-gender} for accuracy (\%).
For formality, as the test set is the same for both formalities, the baseline M-Acc for the two formality labels add up to 1.0.
This is not the case for gender control.

\noindent
\textbf{Quality Evaluation}
We use COMET$\uparrow$~\cite{rei-etal-2020-comet}\footnote{with \texttt{Unbabel/wmt22-comet-da} ($\times$100 for readability)} as the main translation quality metric, 
and additionally report BLEU$\uparrow$\footnote{
using sacreBLEU \cite{post-2018-call} with confidence intervals: bs:1000|rs:12345|c:mixed|e:no|tok:13a|s:exp|v:2.3.1} to compare to prior works.
Note that BLEU is impacted by $n$-gram matches on the correct formality or gendered words, 
while COMET is less susceptible to the artifact.
For COMET score comparisons, we run paired T-tests and bootstrap resampling using \texttt{comet-compare}.
We use "*" or "\dag" to mark systems better or worse than the base pretrained model at $p=0.05$.

\noindent
\textbf{Human Evaluation}
To test the transfer to real low-resource languages,
we conduct a human evaluation on Bengali, which was marked as low-resource in the NLLB-200 training data~\cite{nllb}.
Details on the evaluation are in \autoref{appendix:human_eval_details}. 

\noindent
\textbf{Baselines}
Few existing works experimented on the same data conditions as ours.
An exception is the ``\texttt{mBART-large Gold Finetuned}'' model by~\citet{rippeth-etal-2022-controlling}, 
who finetuned mBART \cite{liu-etal-2020-multilingual-denoising} on parts of CoCoA-MT \cite{nadejde-etal-2022-cocoa} for formality control.
Their results overlap with our 
supervised results on \texttt{en}$\rightarrow$\{de, es, hi\} and 
zero-shot results on \texttt{en}$\rightarrow$\texttt{ru}.
Other than this,
the majority of prior works used \textit{more relaxed} data conditions than ours,
e.g.,
using an existing attribute classifier that covers \textit{zero-shot} languages for pseudo-labeling \cite{lee-etal-2023-improving-formality} or hypothesis reranking \cite{wu-etal-2023-improving}.
We report these results in Appendix \ref{appendix:shared_tasks}.
Overall, our model's performance is comparable to the leading systems.



\section{Supervised Conditions}
\label{subsec:res_supervised}
\autoref{tab:formality_supervised} and \autoref{tab:gender_supervised}  show formality and gender control results respectively with supervised controllers on NLLB-200.
Overall, both finetuning and CG are able to steer the output towards given attributes, while maintaining the original translation quality or at the cost of a slight degradation.

\begin{table}[t!]
\small
\centering
\setlength\tabcolsep{1.3pt} 
\begin{tabular}{llrrrcc}
\toprule
 & 
\textbf{Model} & 
\textbf{F}$_{\text{ormal}}$ & 
\textbf{I}$_{\text{nformal}}$ & 
\textbf{Avg.} & 
\textbf{BLEU} &
\textbf{COMET$_{2022}$}\\ 
\midrule
\multirow{3}{*}{\shortstack[l]{\texttt{en}$\rightarrow$\texttt{de}}}
& base
& 45.6
& 54.4
& $-$
& 35.7{\color{gray!80}\scriptsize$\pm$1.0}
& 82.1\phantom{*} \\
& $+$CG
& 95.0
& 89.6
& 92.3
& 38.4{\color{gray!80}\scriptsize$\pm$1.1}
& 81.6\dag \\
& $+$FT
& 100.0
& 100.0
& 100.0
& 43.6{\color{gray!80}\scriptsize$\pm$1.2}
& 83.8* \\
\multicolumn{2}{r}{\citeauthor{rippeth-etal-2022-controlling}}
& 93.6
& 77.4
& 85.5
& 37.4\phantom{\scriptsize$\pm$0.0}
& $-$
\\
\midrule
\multirow{3}{*}{\shortstack[l]{\texttt{en}$\rightarrow$\texttt{es}}}
& base
& 29.7
& 70.3
& $-$
& 40.0{\color{gray!80}\scriptsize$\pm$1.1}
& 83.9\phantom{*}  \\
& $+$CG
& 72.9
& 92.4
& 82.7 
& 41.2{\color{gray!80}\scriptsize$\pm$1.2}
& 84.4* \\
& $+$FT
& 100.0
& 95.9
& 98.0
& 46.0{\color{gray!80}\scriptsize$\pm$1.2}
& 85.5*\\
\multicolumn{2}{r}{\citeauthor{rippeth-etal-2022-controlling}} 
& 96.7
& 82.7
& 89.7
& 38.3\phantom{\scriptsize$\pm$0.0}
& $-$ 
\\
\midrule
\multirow{3}{*}{\shortstack[l]{\texttt{en}$\rightarrow$\texttt{fr}}}
& base
& 76.8
& 23.2
& $-$
& 36.0{\color{gray!80}\scriptsize$\pm$1.1}
& 80.8\phantom{*}  \\
& $+$CG
& 99.8
& 77.2
& 88.5
& 38.8{\color{gray!80}\scriptsize$\pm$1.2}
& 80.9\phantom{*}  \\
& $+$FT
& 100.0
& 99.3
& 99.7
& 43.0 {\color{gray!80}\scriptsize$\pm$1.1}
& 83.0*\\
\midrule
\multirow{3}{*}{\shortstack[l]{\texttt{en}$\rightarrow$\texttt{hi}}}
& base
& 96.7
& \hspace{4pt}3.3
& $-$
& 24.0{\color{gray!80}\scriptsize$\pm$0.9}
& 75.5\phantom{*}  \\
& $+$CG
& 99.3
& 30.7
& 65.0
& 24.3{\color{gray!80}\scriptsize$\pm$0.9}
& 75.0\dag \\
& $+$FT
& 99.6
& 99.2
& 99.4
& 36.4{\color{gray!80}\scriptsize$\pm$1.0}
& 81.7*\\
\multicolumn{2}{r}{\citeauthor{rippeth-etal-2022-controlling}} 
& 98.5
& 64.7
& 81.6 
& 28.7\phantom{\scriptsize$\pm$0.0}
& $-$
\\
\midrule
\multirow{3}{*}{\shortstack[l]{\texttt{en}$\rightarrow$\texttt{it}}}
& base
& \hspace{4pt}3.2
& 96.8
& $-$
& 41.3{\color{gray!80}\scriptsize$\pm$1.1}
& 84.9\phantom{*}  \\
& $+$CG
& 18.7
& 99.5
& 59.1
& 40.6{\color{gray!80}\scriptsize$\pm$1.1}
& 84.1\dag \\
& $+$FT
& 98.6
& 99.3
& 99.0
& 49.6{\color{gray!80}\scriptsize$\pm$1.1}
& 86.0*\\
\bottomrule
\end{tabular}
\caption{\label{tab:formality_supervised} Formality control results in \textit{supervised} condition (controllers trained on formality-annotated data).}
\end{table}
\begin{table}[t!]
\small
\centering
\setlength\tabcolsep{0.5pt} 
\begin{tabular}{llcccccccccccccc}
\toprule
 & 
\textbf{Model} & 
\textbf{F}$_{\text{eminine}}$ & 
\textbf{M}$_{\text{asculine}}$ & 
\textbf{Global} & 
\textbf{BLEU} &
\textbf{COMET$_{2022}$} &
\\
\midrule
\multirow{3}{*}{\shortstack[l]{\texttt{en}$\rightarrow$\texttt{es}}}
& base
& 58.8
& 86.7
& 73.6
& 45.0{\color{gray!80}\scriptsize$\pm$1.2}
& 84.9\phantom{*}  \\
& $+$CG
& 75.0
& 89.7
& 82.8
& 44.7{\color{gray!80}\scriptsize$\pm$1.2}
& 84.7\phantom{*}  \\
& $+$FT
& 90.2
& 89.7
& 86.9
& 43.7{\color{gray!80}\scriptsize$\pm$1.2}
& 84.0\dag
\\
\bottomrule
\end{tabular}
\caption{\label{tab:gender_supervised} 
Grammatical gender control results in \textit{supervised} condition (cross-domain: controller trained on gender-annotated data from a different domain).}
\end{table}

\begin{table*}[ht!]
\small
\centering
\setlength\tabcolsep{2.7pt} 
\begin{tabular}{llrrrccccccccccc}
\toprule
 &&
 \multicolumn{5}{c}{\textbf{Pretrained Massively Multilingual}} &&
 \multicolumn{5}{c}{\textbf{Transformer-base}} \\
\cmidrule{3-7}  \cmidrule{9-13}
 & 
\textbf{Model} & 
\textbf{F}$_{\text{ormal}}$ & 
\textbf{I}$_{\text{nformal}}$ & 
\textbf{Avg.} & 
\textbf{BLEU} &
\textbf{\textbf{COMET$_{2022}$}}
 && 
\textbf{F}$_{\text{ormal}}$ & 
\textbf{I}$_{\text{nformal}}$ & 
\textbf{Avg.} & 
\textbf{BLEU} &
\textbf{COMET$_{2022}$}\\ 
\midrule
\multirow{8}{*}{\shortstack[l]{\texttt{en}$\rightarrow$\texttt{pt}}}
& base
& 47.7
& 52.3 
& $-$
& 41.7{\color{gray!80}\scriptsize$\pm$1.1}
& 85.1\phantom{*} 
&
& 35.8
& 64.2 
& $-$ 
& 38.7{\color{gray!80}\scriptsize$\pm$1.1}
& 82.2\phantom{*} 
\\
& $+$CG (\texttt{de})
& 75.6
& 74.2
& 74.9
& 43.0{\color{gray!80}\scriptsize$\pm$1.1}
& 84.9\phantom{*} 
& 
& 50.0
& 72.8
& 61.4
& 38.7{\color{gray!80}\scriptsize$\pm$1.1}
& 81.8\dag
\\
& $+$FT (\texttt{de})
& 99.0
& 45.5
& 72.3
& 40.4{\color{gray!80}\scriptsize$\pm$1.0}
& 85.3\phantom{*} 
&
& 79.2
& 71.2
& 75.2
& 39.6{\color{gray!80}\scriptsize$\pm$1.1}
& 82.7*
\\
& $+$CG (\texttt{es})
& 85.4
& 83.6
& \underline{84.5}
& 43.8{\color{gray!80}\scriptsize$\pm$1.0}
& 85.0\phantom{*} 
&
& 53.3
& 79.8
& 66.6
& 38.8{\color{gray!80}\scriptsize$\pm$1.1}
& 81.7\dag
\\
& $+$FT (\texttt{es})
& 99.8
& 28.7
& 64.3
& 40.3{\color{gray!80}\scriptsize$\pm$1.0}
& 85.2\phantom{*} 
&
& 93.9
& 80.1 
& 87.0 
& 40.5{\color{gray!80}\scriptsize$\pm$1.0}
& 82.5*
\\
& $+$CG (\texttt{multi})
& 84.8
& 80.0
& 82.4
& 43.7{\color{gray!80}\scriptsize$\pm$1.1}
& 84.9\phantom{*} 
& 
& 55.9
& 80.8
& 68.4
& 39.0{\color{gray!80}\scriptsize$\pm$1.1}
& 81.8\dag
\\
& $+$FT (\texttt{multi})
& 99.5
& 51.0
& 75.3
& 42.3{\color{gray!80}\scriptsize$\pm$1.0}
& 85.9{*}
&
& 95.8
& 81.9
& \underline{88.9}
& 41.4{\color{gray!80}\scriptsize$\pm$1.0}
& 83.1*
\\
& $+$CG $+$FT (\texttt{multi})
& 100.0
& 83.2
& \textbf{91.6}
& 42.1{\color{gray!80}\scriptsize$\pm$1.0}
& 85.7*
&
& 97.8
& 93.7
& \textbf{95.8}
& 41.0{\color{gray!80}\scriptsize$\pm$1.0}
& 82.4\phantom{*} 
\\
\midrule
\multirow{8}{*}{\shortstack[l]{\texttt{en}$\rightarrow$\texttt{ru}}}
& base
& 55.0
& 45.0 
& $-$
& 30.3{\color{gray!80}\scriptsize$\pm$1.0}
& 83.7\phantom{*} 
&
& 43.9
& 56.1 
& $-$ 
&
24.2{\color{gray!80}\scriptsize$\pm$1.0}
& 75.9\phantom{*} 
\\
& $+$CG (\texttt{de})
& 87.3
& 77.7 
& 82.5
& 32.2{\color{gray!80}\scriptsize$\pm$1.0}
& 83.1\phantom{*} 
& 
& 67.2
& 71.8
& 69.5
& 24.6{\color{gray!80}\scriptsize$\pm$0.9}
& 75.0\dag
\\
& $+$FT (\texttt{de})
& 99.5
& 84.7
& \underline{92.1}
& 33.0{\color{gray!80}\scriptsize$\pm$1.1}
& 84.2*  
&
& 84.0
& 69.3
& 76.7
& 25.0{\color{gray!80}\scriptsize$\pm$1.0}
& 75.8\phantom{*} 
\\
& $+$CG (\texttt{es})
& 86.8
& 73.9
& 80.4
& 32.4{\color{gray!80}\scriptsize$\pm$1.0}
& 83.2\phantom{*} 
&
& 61.7
& 76.8
& 69.5
& 24.8{\color{gray!80}\scriptsize$\pm$1.0}
& 75.0\dag
\\
& $+$FT (\texttt{es})
& 98.3
& 60.6
& 79.5
& 32.8{\color{gray!80}\scriptsize$\pm$1.1}
& 84.1*   
&
& 83.5
& 68.6
& 76.1
& 26.1{\color{gray!80}\scriptsize$\pm$1.0}
& 76.6*
\\
& $+$CG (\texttt{multi})
& 87.3
& 78.2
& 82.8
& 32.2{\color{gray!80}\scriptsize$\pm$1.0}
& 83.2\phantom{*} 
&
& 72.2
& 80.9
& 76.6
& 25.0{\color{gray!80}\scriptsize$\pm$1.0}
& 75.0\dag
\\
& $+$FT (\texttt{multi})
& 99.8
& 79.6
& 89.7
& 33.0{\color{gray!80}\scriptsize$\pm$1.1}
& 84.2*
&
& 87.5
& 69.8
& \underline{78.7}
& 25.9{\color{gray!80}\scriptsize$\pm$1.0}
& 77.0*
\\
& $+$CG $+$FT (\texttt{multi})
& 100.0
& 93.0 
& \textbf{96.5}
& 33.1{\color{gray!80}\scriptsize$\pm$1.0}
& 84.4*
&
& 96.2
& 91.3
& \textbf{93.8}
& 26.2{\color{gray!80}\scriptsize$\pm$1.0}
& 76.2\phantom{*} 
\\
&
\citet{rippeth-etal-2022-controlling}
& 100.0
& 13.8
& 56.9
& 23.5\phantom{{\scriptsize$\pm$0.0}}
& $-$
&
& $-$
& $-$
& $-$
& $-$
& $-$
\\
\midrule
\multirow{10}{*}{\shortstack[l]{\texttt{en}$\rightarrow$\texttt{ko}}}
& base
& 50.9
& 49.1
& $-$
& 15.7{\color{gray!80}\scriptsize$\pm$0.7}
& 82.6\phantom{*} 
&  
& 32.0
& 68.0
& $-$
& 10.6{\color{gray!80}\scriptsize$\pm$0.6}
& 74.0\phantom{*} 
\\
& $+$CG (\texttt{de})
& 67.0
& 64.6
& \underline{65.8}
& 15.7{\color{gray!80}\scriptsize$\pm$0.7}
& 82.1\dag
& 
& 45.2 
& 78.2
& 61.7
& 10.4{\color{gray!80}\scriptsize$\pm$0.6}
& 73.4\dag
\\
& $+$FT (\texttt{de})
& 67.8
& 54.2
& 61.0
& 12.8{\color{gray!80}\scriptsize$\pm$0.6}
& 84.1*
&  
& 42.7
& 66.4
& 54.6
& 10.7{\color{gray!80}\scriptsize$\pm$0.6}
& 74.0\phantom{*} 
\\
& $+$CG (\texttt{es})
& 68.9
& 61.6
& 65.3
& 15.1{\color{gray!80}\scriptsize$\pm$0.8}
& 82.1\dag
& 
& 46.3
& 77.6
& 62.0
& 10.7{\color{gray!80}\scriptsize$\pm$0.6}
& 74.1\phantom{*} 
\\
& $+$FT (\texttt{es})
& 64.4
& 47.3
& 55.9
& 14.0{\color{gray!80}\scriptsize$\pm$0.7}
& 84.4* 
&  
& 47.4
& 62.7
& 55.1
& 11.7{\color{gray!80}\scriptsize$\pm$0.6}
& 75.2*
\\
& $+$CG (\texttt{multi})
& 67.0
& 61.7
& 64.4
& 15.5{\color{gray!80}\scriptsize$\pm$0.8}
& 82.2\phantom{*} 
& 
& 46.0
& 78.1
& \underline{62.1}
& 10.6{\color{gray!80}\scriptsize$\pm$0.6}
& 74.1\phantom{*} 
\\
& $+$FT (\texttt{multi})
& 68.5
& 46.2
& 57.4
& 13.4{\color{gray!80}\scriptsize$\pm$0.7}
& 84.7*
&  
& 48.3
& 68.4
& 58.4
& 11.0{\color{gray!80}\scriptsize$\pm$0.6}
& 74.4\phantom{*} 
\\
& $+$CG $+$FT (\texttt{multi})
& 70.0
& 63.5
& \textbf{66.8} 
& 13.2{\color{gray!80}\scriptsize$\pm$0.7}
&
84.2*
&
& 58.9
& 81.8
& \textbf{70.4}
& 10.8{\color{gray!80}\scriptsize$\pm$0.6}
& 73.4\dag
\\
\cmidrule{2-13}
& $+$oracle CG (\texttt{ko})
& 70.3
& 62.6
& 66.5
& 15.2{\color{gray!80}\scriptsize$\pm$0.7}
& 81.7\dag 
& 
& 58.9
& 82.3
& 70.6
&11.2{\color{gray!80}\scriptsize$\pm$0.6}
&74.5*
\\
& $+$oracle FT (\texttt{ko})
& 79.4
& 93.5
& 86.5
& 22.2{\color{gray!80}\scriptsize$\pm$0.9}
& 86.2*
&
& 86.7
& 97.9
& 92.3
& 19.1{\color{gray!80}\scriptsize$\pm$0.9}
& 74.0*
\\
\bottomrule
\end{tabular}
\caption{\label{tab:formality_tgt_zs} 
Zero-shot formality control results.
\textbf{Best} and \underline{second best} results under the same data condition are marked.}
\end{table*}

\paragraph{Finetuning more effective than classifier guidance in supervised conditions:} 
A comparison of scores in \autoref{tab:formality_supervised} and \autoref{tab:gender_supervised} clearly shows FT is more effective than CG. 
For formality control, FT consistently scores nearly 100\% M-Acc.
It also substantially improves the quality scores due to adapting towards the specific domain of the attribute-annotated data,
which is the same as the test domain in this case.
On the other hand for CG, 
while it also improves the formality accuracy, 
the scores lag behind finetuning in both accuracy and quality.
The gap is especially prominent on \texttt{hi} and \texttt{it}, where the underlying NLLB model has a strong bias towards a single formality: 
the accuracy for the rare formality is nearly zero (3.3\% and 3.2\% respectively). 
This is likely to do with NLLB's training data, which might be skewed towards one single formality for some languages.
In this case, CG can only partly recover the ability to generate translation in the formality NLLB is unfamiliar with.
These results indicate that CG is only effective when the underlying model does not suffer from an absolute bias towards one attribute.

\paragraph{Classifier guidance more robust to domain mismatch:}
As motivated in \S\ref{subsec:gender_data},
the gender control results in \autoref{tab:gender_supervised} allow us to assess the impact of domain mismatch between the controller training data and the test data, a very realistic scenario in practice.
Here, while finetuning achieves higher accuracy for gendered words, it also degrades translation quality by 0.9 COMET.
This provides further evidence that the previously improved COMET scores (\autoref{tab:formality_supervised}) are results of finetuning on in-domain data.
In contrast, the translation quality with CG does not significantly differ from NLLB by the T-tests, suggesting its stronger domain robustness.
We hypothesize it is because CG operates on the last decoder layer's hidden states, which are just one projection away from the output vocabulary.
These representations likely contain more word-level than domain information,
which is precisely needed in the task of attribute control.

\section{Zero-Shot Conditions}
\subsection{New Target Languages} \label{subsec:new_target}
Now we transfer the trained controllers to target languages unseen when training the controllers, 
i.e., those without attribute annotation.
In \autoref{tab:formality_tgt_zs} and \autoref{tab:gender_control}, we report the results on formality and gender control respectively.
In \autoref{tab:formality_tgt_zs}, we also compare the single-direction and multilingual controllers as motivated in \S\ref{subsec:data_formality}.

\paragraph{Gap between finetuning and classifier guidance shrinks in zero-shot conditions:}
While finetuning was consistently leading in supervised conditions (\S\ref{subsec:res_supervised}), 
now under zero-shot conditions with unseen target languages, the gap shrinks.
For formality control, on Korean, the most distant language,
CG consistently achieves stronger control results than finetuning, 
indicating more robustness when transferring to unfamiliar settings. 
Overall in \autoref{tab:formality_tgt_zs},
for the main experiments on NLLB-200, 
CG outperforms FT in 7 of the 9 pairwise comparisons (\{\texttt{de}, \texttt{es}, \texttt{multi}\} $\times$ 3 target languages).
With gender control results in \autoref{tab:gender_control}, 
finetuning achieves stronger control accuracy (avg. $+$4.9\% abs.) 
but degrades translation quality ($-$0.6 COMET) due to domain mismatch.
On the other hand, CG retains the translation quality. 
This confirms the previous finding (\S\ref{subsec:res_supervised}) on its stronger domain robustness.

\paragraph{Multilingual controllers help when the base model is not massively multilingual:}
In \autoref{tab:formality_tgt_zs}, controllers trained on multiple translation directions (\texttt{multi}) are compared to those trained on single directions (\texttt{en}$\rightarrow$\texttt{es} or \texttt{de}).
On Transformer-base, \texttt{multi} consistently outperforms its single-direction counterparts, regardless whether the controller is finetuning- or CG-based.
In contrast, 
for the pretrained NLLB, 
there is no clear distinction between the multilingual systems and rest.
This indicates that NLLB does not further benefit from multilinguality in the controller training stage, 
likely because it already underwent a massively multilingual pretraining stage.
This shows that massively multilingual models are a useful basis for attribute control especially when annotated resources are limited to single languages.

\begin{table}[t!]
\small
\centering
\setlength\tabcolsep{0pt} 
\begin{tabular}{llcccccccccccccc}
\toprule
 & 
\textbf{Model} & 
\textbf{F}$_{\text{eminine}}$ & 
\textbf{M}$_{\text{asculine}}$ & 
\textbf{Global} & 
\textbf{BLEU} &
\textbf{COMET} 
\\ 
\midrule
\multirow{4}{*}{\shortstack[l]{\texttt{en}$\rightarrow$\texttt{it}}}
& base
& 53.8
& 88.9
& 73.1
& 35.1{\color{gray!80}\scriptsize$\pm$1.0}
& 84.1\phantom{*} 
\\
& $+$CG
& 72.3
& 92.8
& 83.6
& 35.4{\color{gray!80}\scriptsize$\pm$1.1}
& 83.7\phantom{*} 
\\
& $+$FT
& 83.6
& 91.2
& 87.8
& 34.4{\color{gray!80}\scriptsize$\pm$1.0}
& 83.5\dag
\\
& $+$CG $+$FT  
& 88.6
& 94.5
& \textbf{91.8}
&33.4{\color{gray!80}\scriptsize$\pm$1.0}
& 82.6\dag
\\
\midrule
\multirow{4}{*}{\shortstack[l]{\texttt{en}$\rightarrow$\texttt{fr}}}
& base 
& 55.3
& 88.4
& 72.4
& 38.3{\color{gray!80}\scriptsize$\pm$1.3}
& 82.6\phantom{*} 
\\
& $+$CG
& 67.8
& 90.3
& 79.4
& 38.7{\color{gray!80}\scriptsize$\pm$1.2}
& 82.5\phantom{*} 
\\
& $+$FT
& 78.9
& 90.8
& 85.0
& 38.2{\color{gray!80}\scriptsize$\pm$1.2}
& 82.0\dag
\\
& $+$CG $+$FT  
& 87.0
& 91.9
& \textbf{89.5}
& 37.4{\color{gray!80}\scriptsize$\pm$1.2}
& 81.9\dag
\\
\bottomrule
\end{tabular}
\caption{\label{tab:gender_control} 
Zero-shot grammatical gender control results on \textit{new target} languages with \textit{domain mismatch}.}
\end{table}
\begin{table}[t]
\small
\centering
\setlength\tabcolsep{0.5pt} 
\begin{tabular}{llcccccccccccccc}
\toprule
&
\textbf{Model} & 
\textbf{Quality} & 
\textbf{Formality} & 
\textbf{Win} & 
\textbf{Win \& Tie} &
\\ 
&
&
(1-5)
& 
(1-3)
& 
(\%)
& 
(\%)
\\
\midrule
$(1)$
&
NLLB-200
& 
4.25{\color{gray!80}\scriptsize$\pm$0.75}
&
2.69{\color{gray!80}\scriptsize$\pm$0.46}
&
$-$
&
$-$
\\
$(2)$
&
CG (\texttt{multi}) formal
& 4.00{\color{gray!80}\scriptsize$\pm$0.79}
& 2.63{\color{gray!80}\scriptsize$\pm$0.48}
& 56.3
& 81.3
\\
$(3)$
&
CG (\texttt{multi}) inf.
& 4.44{\color{gray!80}\scriptsize$\pm$0.70}
& 2.38{\color{gray!80}\scriptsize$\pm$0.69}
& 62.5
& 93.8
\\
$(4)$
&
FT (\texttt{multi}) formal
& 
4.31{\color{gray!80}\scriptsize$\pm$0.85}
&
2.63{\color{gray!80}\scriptsize$\pm$0.48}
&
43.8
&
68.8
\\
$(5)$
&
FT (\texttt{multi}) inf.
& 4.13{\color{gray!80}\scriptsize$\pm$1.05}
& 2.44{\color{gray!80}\scriptsize$\pm$0.49}
& 62.5
& 93.8
\\

\bottomrule
\end{tabular}
\caption{\label{tab:human_eval_bn} 
Human evaluation on Bengali,
with quality on a 5-point scale$\uparrow$
and formality on a 3-point scale ($\uparrow$: formal) with {\color{gray!80}standard deviations}.
Last two columns show pairwise comparison of formality scores to baseline NLLB-200 given the same source sentences (winning: scoring more in the direction of the desired formality).
}
\end{table}
\begin{table*}[ht!]
\small
\centering
\setlength\tabcolsep{5pt} 
\begin{tabular}{llrrrcccrrrcc}
\toprule
 &&
 \multicolumn{5}{c}{\textbf{Source\colorbox{gray!30}{Formal}}} &&
 \multicolumn{5}{c}{\textbf{Source\colorbox{blue!15}{Informal}}} \\
 \cmidrule{3-7}  \cmidrule{9-13}
 & 
\textbf{Model} & 
\textbf{F}$_{\text{ormal}}$ & 
\textbf{I}$_{\text{nformal}}$ & 
\textbf{Avg.} & 
\textbf{BLEU} &
\textbf{COMET$_{2022}$} &&
\textbf{F}$_{\text{ormal}}$ & 
\textbf{I}$_{\text{nformal}}$ & 
\textbf{Avg.} & 
\textbf{BLEU} &
\textbf{COMET$_{2022}$}\\ 
\midrule
\multirow{1}{*}{\shortstack[l]{\texttt{X}$\rightarrow$\texttt{de}}}
& base
&\cellcolor[HTML]{d3d1d1}77.8
& 22.2
& $-$ 
& 23.9{\color{gray!80}\scriptsize$\pm$0.5}
& 79.0\phantom{*}  
&
& 48.5
& \cellcolor{blue!15}51.5
& $-$
& 24.7{\color{gray!80}\scriptsize$\pm$0.5}
& 79.3\phantom{*} 
\\
& $+$CG
&\cellcolor[HTML]{d3d1d1}98.6
& 71.5
& 85.1
& 25.9{\color{gray!80}\scriptsize$\pm$0.5}
& 78.7\phantom{*} 
&
& 94.0
& \cellcolor{blue!15}87.7
& 90.9
& 27.0{\color{gray!80}\scriptsize$\pm$0.5}
& 79.0\phantom{*}
\\
& $+$FT 
& 100.0\cellcolor[HTML]{d3d1d1}
& 100.0
& 100.0
& 30.1{\color{gray!80}\scriptsize$\pm$0.7}
& 80.7*
&
& 100.0
& \cellcolor{blue!15} 99.7
& 99.9
& 30.0{\color{gray!80}\scriptsize$\pm$0.6}
& 80.7*
\\
\texttt{X}$\rightarrow$\texttt{es}
& base
&\cellcolor[HTML]{d3d1d1}57.8
& 42.2
& $-$
& 29.8{\color{gray!80}\scriptsize$\pm$0.5}
& 82.7\phantom{*}
&
& 20.3
& \cellcolor{blue!15}79.7
& $-$
& 29.9{\color{gray!80}\scriptsize$\pm$0.5}
& 82.7\phantom{*}
\\
& $+$CG
& \cellcolor[HTML]{d3d1d1}86.7
& 73.3
& 80.0
& 30.5{\color{gray!80}\scriptsize$\pm$0.8}
& 82.3\phantom{*} 
&
& 67.5
& \cellcolor{blue!15}93.7
& 80.6
& 31.1{\color{gray!80}\scriptsize$\pm$0.6}
& 82.3\phantom{*} 
\\
& $+$FT 
&\cellcolor[HTML]{d3d1d1}99.6
& 77.4
& 88.5
& 32.8{\color{gray!80}\scriptsize$\pm$0.7}
& 83.9*
&
& 99.8
& \cellcolor{blue!15}97.8
& 98.8
& 33.2{\color{gray!80}\scriptsize$\pm$0.7}
& 83.9*
\\
\multirow{1}{*}{\shortstack[l]{\texttt{X}$\rightarrow$\texttt{fr}}}
& base
&\cellcolor[HTML]{d3d1d1}97.0
& \hspace{4pt}3.0
& $-$
& 29.5{\color{gray!80}\scriptsize$\pm$0.6}
& 79.1\phantom{*}  
&
& 87.7
& \cellcolor{blue!15}12.3
& $-$ 
& 30.3{\color{gray!80}\scriptsize$\pm$0.6}
& 79.6\phantom{*} 
\\
& $+$CG
&\cellcolor[HTML]{d3d1d1}99.8
& 40.4
& 70.1
& 30.6{\color{gray!80}\scriptsize$\pm$0.6}
& 78.9\phantom{*}  
&
& 99.9
& \cellcolor{blue!15}59.5
& 79.7
& 32.5{\color{gray!80}\scriptsize$\pm$0.6}
& 79.6\phantom{*} 
\\
& $+$FT 
& \cellcolor[HTML]{d3d1d1}99.9
& 99.4
& 99.7
& 34.2{\color{gray!80}\scriptsize$\pm$0.7}
& 81.0*
&
& 100.0
& \cellcolor{blue!15} 100.0
& 100.0
& 35.6{\color{gray!80}\scriptsize$\pm$0.6}
& 81.5*
\\
\multirow{1}{*}{\shortstack[l]{\texttt{X}$\rightarrow$\texttt{hi}}}
& base
&\cellcolor[HTML]{d3d1d1}98.2
& \hspace{4pt}1.8
& $-$
& 20.2{\color{gray!80}\scriptsize$\pm$0.4}
& 73.2\phantom{*} 
&
& 98.4
& \hspace{4pt}\cellcolor{blue!15}1.6
& $-$
& 20.8{\color{gray!80}\scriptsize$\pm$0.4}
& 73.6\phantom{*} 
\\
& $+$CG
&\cellcolor[HTML]{d3d1d1}99.2
& \hspace{4pt}9.8
& 54.5
& 20.3{\color{gray!80}\scriptsize$\pm$0.4}
& 73.0\phantom{*} 
&
& 99.2
& \cellcolor{blue!15}12.3
& 55.7
& 20.8{\color{gray!80}\scriptsize$\pm$0.4}
& 73.4\phantom{*} 
\\
& $+$FT 
& \cellcolor[HTML]{d3d1d1}99.4
& 99.3
& 99.4
& 26.5{\color{gray!80}\scriptsize$\pm$0.6}
& 75.3*
&
& 99.7
& \cellcolor{blue!15} 99.5
& 99.6
& 27.7{\color{gray!80}\scriptsize$\pm$0.6}
& 75.8*
\\
\multirow{1}{*}{\shortstack[l]{\texttt{X}$\rightarrow$\texttt{it}}}
& base
&\cellcolor[HTML]{d3d1d1}23.0
& 77.0
& $-$
& 27.6{\color{gray!80}\scriptsize$\pm$0.6}
& 83.5\phantom{*} 
&
& \hspace{4pt}1.5
& \cellcolor{blue!15}98.5
& $-$
& 28.0{\color{gray!80}\scriptsize$\pm$0.6}
& 83.6\phantom{*} 
\\
& $+$CG
&\cellcolor[HTML]{d3d1d1}45.8
& 88.1
& 67.0
& 28.1{\color{gray!80}\scriptsize$\pm$0.6}
& 82.9\phantom{*} 
&
& 17.1
& \cellcolor{blue!15}99.4
& 58.3
& 28.0{\color{gray!80}\scriptsize$\pm$0.6}
& 82.9\phantom{*} 
\\
& $+$FT 
& \cellcolor[HTML]{d3d1d1} 99.2
& 88.1
& 93.7
& 32.4{\color{gray!80}\scriptsize$\pm$0.7}
& 84.4*
&
& 98.2
& 99.2\cellcolor{blue!15}
& 98.7
& 32.8{\color{gray!80}\scriptsize$\pm$0.7}
& 84.5*
\\
\bottomrule
\end{tabular}
\caption{\label{tab:formality_src_zs} 
Zero-shot formality control results on \textit{new source} languages, using controllers trained on English as source.
Sources are \{de, es, fr, hi, it\}.
Colored columns indicate source formality agreeing with desired target formality.
}
\end{table*}

\begin{table}[t]
\small
\centering
\setlength\tabcolsep{0.5pt} 
\begin{tabular}{llcccccccccccccc}
\toprule
 & 
\textbf{Model} & 
\textbf{F}$_{\text{eminine}}$ & 
\textbf{M}$_{\text{asculine}}$ & 
\textbf{Global} & 
\textbf{BLEU} &
\textbf{COMET$_{2022}$} &
\\ 
\midrule
\multirow{3}{*}{\shortstack[l]{\texttt{es}$\rightarrow$\texttt{it}}}
& base
& 79.4
& 89.3
& 85.2
& 30.0{\color{gray!80}\scriptsize$\pm$1.5}
& 83.3\phantom{*} \\
& $+$CG
& 87.6
& 92.3
& 90.4 
& 29.5{\color{gray!80}\scriptsize$\pm$1.4}
& 82.9\dag \\
& $+$FT
& 90.9
& 90.5
& 90.7
& 30.0{\color{gray!80}\scriptsize$\pm$1.3}
& 82.9\dag\\
\midrule
\multirow{3}{*}{\shortstack[l]{\texttt{fr}$\rightarrow$\texttt{it}}}
& base
& 75.4
& 90.4
& 84.2
& 28.1{\color{gray!80}\scriptsize$\pm$1.4}
& 82.6\phantom{*}\\
& $+$CG
& 85.1
& 94.1
& 90.4 
& 27.7{\color{gray!80}\scriptsize$\pm$1.4}
& 82.3\phantom{*} \\
& $+$FT
& 90.4
& 93.6
& 92.3
&28.6{\color{gray!80}\scriptsize$\pm$1.4}
& 82.5\phantom{*}\\
\midrule
\multirow{3}{*}{\shortstack[l]{\texttt{es}$\rightarrow$\texttt{fr}}}
& base
& 83.2
& 87.0
& 85.3
& 31.2{\color{gray!80}\scriptsize$\pm$1.4}
& 79.9\phantom{*} \\
& $+$CG
& 86.8
& 88.8
& 87.9 
& 31.3{\color{gray!80}\scriptsize$\pm$1.4}
& 79.8\phantom{*} \\
& $+$FT
& 89.2
& 88.5
& 88.8
& 31.4{\color{gray!80}\scriptsize$\pm$1.5}
& 79.7\phantom{*} \\
\midrule
\multirow{3}{*}{\shortstack[l]{\texttt{it}$\rightarrow$\texttt{fr}}}
& base
& 76.1
& 87.2
& 84.3
& 31.5{\color{gray!80}\scriptsize$\pm$1.3}
& 80.4\phantom{*} \\
& $+$CG
& 86.4
& 89.1
& 87.9
& 31.5{\color{gray!80}\scriptsize$\pm$1.3}
& 80.5\phantom{*} \\
& $+$FT
& 90.6
& 88.9
& 89.6
& 31.8{\color{gray!80}\scriptsize$\pm$1.4}
& 80.4\phantom{*}\\
\bottomrule
\end{tabular}
\caption{\label{tab:gender_zs_src_tgt} 
Zero-shot grammatical gender control results on \textit{new source and target} languages.}
\end{table}
\paragraph{Classifier guidance is complementary with finetuning:}
When applying CG on top of the finetuned models, we see the \textit{strongest} control accuracy for both formality and gender control.
This observation is consistent whether the base model is the pretrained NLLB or the normal Transformer-base.
Compared to finetuning alone, the addition of CG also does not degrade translation quality on NLLB.
On the more challenging case of gender control which involves domain mismatch, 
adding CG to finetuning does not impact translation quality on \texttt{fr} and causes a slight degradation on \texttt{it}.
This is likely linked to poor hyperparameter choices in CG: 
due to time constraints we directly used the hyperparameters when applying CG alone,
which are too strong for models already finetuned for attribute control.
We are optimistic for improved scores under more fitting hyperparameters.

\paragraph{Finetuning did not erase knowledge on other languages:}
To our surprise and different from results in domain adaptation \cite{cooper-stickland-etal-2021-multilingual,vu-etal-2022-domains},
finetuning did not erase the pretrained model's knowledge on the target languages absent in supervised finetuning,
as reflected by the translation quality scores (\autoref{tab:formality_tgt_zs}, \ref{tab:gender_control}).
This is not specific to NLLB, 
but also observed on the Transformer-base trained with random initialization on a few translation directions. 
Therefore, 
this phenomenon is not a result of massively multilingual pretraining, 
but more likely linked to the light finetuning strength with limited number of updates and small learning rates.

\paragraph{Comparison to oracle data condition:}
In the bottom rows of \autoref{tab:formality_tgt_zs}, we report the oracle performance of using 400 sentences as supervised data for training the controllers.
Our strongest zero-shot results match the performance of oracle CG, but still lag far behind the upper-bound of finetuning on in-domain data with attribute annotation (oracle FT).
We believe this gap is magnified as Korean is not only linguistically distant from the languages used in training, it also differs in the notion of formality: 
Korean involves multiple levels of formality instead of a binary informal-formal distinction.
For the zero-shot transfer, this means transferring a controller trained for binary control to a multi-class problem with an unknown class mapping, which is naturally more challenging.

\paragraph{Human Evaluation on Bengali:} \label{subsec:human_eval_results}
The results are in 
\autoref{tab:human_eval_bn}.
First, adding attribute control does not appear to impact translation quality.
Second, pairwise comparisons with the baseline show both CG and finetuning are effective in formality control, where CG has slightly higher win ratio than FT against the baseline.
Third, the impact on formality scores is more prominent when steering towards informal translation. 
This likely because the baseline translations already have a high level of formality.
Moreover,
the rare usage of the lowest formality level in Bengali (Appendix \ref{appendix:human_eval_details}) could explain the relatively high formality scores for the systems steered towards ``informal'' (rows $(3)$ and $(5)$).

\subsection{New Source and Target Languages} \label{subsec:new_source_target}

\paragraph{New source languages easier than new target languages:}
In \autoref{tab:formality_src_zs}, we report the results of transferring controllers trained with English source to new source languages.
Contrasting these scores with the target-side zero-shot results in \autoref{tab:formality_tgt_zs}, 
it is clear that transferring to new source languages is a much easier task.
This is expected, as attribute-controlled translation primarily places lexical constraints on the target side.
Once the controller can generate translations with the correct attribute, swapping the source language does not pose a large challenge.
Even when the source formality disagrees with the desired output formality (uncolored columns in \autoref{tab:formality_src_zs}),
the controllers are able to steer the translations toward the required attributes.

\paragraph{NLLB struggles to preserve source attributes:}
Contrasting the colored ``base'' cell in \autoref{tab:formality_src_zs} with its uncolored counterpart,  
we see that NLLB does have some notion of formality in the source sentences, 
as source sentences with the correct formality improves accuracy on the desired formality (57.8\ vs. 42.2\% and 79.7\ vs. 20.3\%).
However, 
the signals in the input alone are insufficient for generating the correct formality.
This is confirmed by another zero-shot experiment when both the source and target languages are new (\autoref{tab:gender_zs_src_tgt}).
Here the sources already contain the correct grammatical genders.
Despite this, NLLB cannot fully utilize  the signals in the source, especially on the feminine gender.
Its accuracy (76.1-83.2\%) still lags behind the masculine class (87.0-90.4\%).
Both CG and finetuning substantially improve the accuracy and mostly close the gap between the two grammatical classes.
This shows both approaches strengthen the source signals that are otherwise neglected.

\section{Conclusion}
To generalize attribute-controlled translation to data-scarce conditions, 
we asked the question ``how \textit{transferable} are attribute controllers on pretrained multilingual translation model?’’.
We use a novel classifier guidance method to extend a pretrained NLLB-200 model for attribute control and contrast its performance to finetuning-based control.

Our results led to the following recommendations for upgrading existing multilingual translation systems with attribute control capabilities: \textbf{1)} Given in-domain target sentences annotated with attributes, even as few as the lower hundreds, finetuning is the primary choice. 
\textbf{2)} 
In case of \textit{distant} new target languages or strong \textit{domain} mismatches between the attribute-annotated data and test data, decoding with classifier guidance is more promising. Otherwise finetuning is recommended.
\textbf{3)} In case specific resource constraints preclude finetuning or hosting multiple specialized variants of the underlying model, we then recommend inference-time control by classifier guidance.
\textbf{4)} In case the underlying translation model is not massively multilingual, finetuning the model or training the controller on multiple target languages is beneficial.

\section*{Limitations}
\paragraph{More Fine-Grained Attributes}
Our classifier guidance approach works with discrete labels, 
making it not directly applicable to use-cases with more fine-grained or continuous attributes.
In particular, although the gender classifier training incdlues a gender-neutral class, 
in evaluation we were only able to test two genders, 
limited by the availability of test data.
As more test datasets with fine-grained attributes become available,
our approach can be further improved and validated for these use-cases.

\paragraph{Inference Speed}
Decoding speed is a main downside of our classifier guidance approach. 
This is a result of multiple gradient-based updates of model activations at each decoding time step.
Despite the promising zero-shot results, further speed-up is necessary is make it realistic for deployed systems.

\paragraph{Acknowledgement}
We thank Supriti Sinhamahapatra for help with the human evaluation, and the anonymous reviewers for helpful feedback.
This work was performed on the HoreKa supercomputer funded by the
Ministry of Science, Research and the Arts Baden-Württemberg and by
the Federal Ministry of Education and Research.
Part of this work was supported by funding from the pilot program Core-Informatics of the Helmholtz Association (HGF).

\bibliography{anthology,custom}

\appendix
\section{Dataset Statistics}
\subsection{OPUS-100 Data for Transformer-Base} \label{appendix:opus_data}
The data overview is in \autoref{tab:opus_overview}.
For tokenization, we use the SentencePiece~\cite{kudo-richardson-2018-sentencepiece} model from NLLB-200\footnote{\url{https://github.com/facebookresearch/fairseq/tree/nllb/\#preparing-datasets-for-training}}~\cite{nllb}.
The model is trained to translate from and into English.

\begin{table}[ht!]
\small
\centering
\setlength\tabcolsep{5pt} 
\begin{tabular}{lrrr}
\toprule
\textbf{Direction} & 
\textbf{\# Sentences} & 
\textbf{\# Tokens (en)} &
\textbf{\# Tokens (X)} \\
\midrule
en-es & 1,000,000 & 15,482,094 & 16,422,413\\
en-de & 1,000,000 & 17,952,717 & 20,142,507\\
en-fr & 1,000,000 & 21,495,343 & 26,634,530\\
en-hi &  534,319  &  8,723,899 & 10,913,496\\
en-it & 1,000,000 &  14,435,382 & 15,524,589\\
en-ko & 1,000,000 & 11,290,102 & 9,552,148\\
en-pt & 1,000,000 & 13,879,742 & 14,410,909\\
en-ru & 1,000,000 & 16,638,782 & 19,630,699 \\
\bottomrule
\end{tabular}
\caption{\label{tab:opus_overview} Overview of OPUS-100 data we used to train the Transformer-base.}
\vspace{-5pt}
\end{table}

\subsection{Details on Domain Mismatch Data} \label{appendix:data_domain_mismatch}
For the grammatical gender control experiments with domain mismatch (\S\ref{subsec:gender_data}),
the training domain differs from the test sets in both style and length.
An overview is shown in \autoref{tab:data_domain_mismatch}.

During training, 
an example tuple of (input, output, attribute label) is:
("the actor finished her work.", "La actriz terminó su trabajo.", 0: feminine)
("the actor finished his work.", "El actor terminó su trabajo.", 1: masculine).
The training sentences are all artificial sentences following this simple subject-verb-objective structure.
This differs significantly from the test sets with public speaking texts.

\begin{table}[ht!]
\small
\centering
\setlength\tabcolsep{0.5pt} 
\begin{tabular}{lccc}
\toprule
\textbf{Split} & 
\textbf{Style} &
\multirow{2}{*}{\shortstack[c]{\textbf{Avg. \# output}\\ \textbf{words per sent.}}}
\\
\\
\midrule
Train & artificial sentences & {\phantom0}5.5 \\ 
Test (supervised) & TED talks & 25.4\\
Test (new tgt lang.) & TED talks & 25.2 \\ 
Test (new src \& tgt lang.) & TED talks & 26.2 \\ 
\bottomrule
\end{tabular}
\caption{\label{tab:data_domain_mismatch} Details on domain mismatch training setup.}
\vspace{-5pt}
\end{table}

\section{Training and Inference Details}\label{appendix:training_inference}
We implemented our approaches in \textsc{Fairseq}~\cite{ott-etal-2019-fairseq} at \url{https://github.com/dannigt/attribute-controller-transfer}.

\subsection{Inference}
\paragraph{Preprocessing}
For CoCoA-MT~\cite{nadejde-etal-2022-cocoa}, 
many test inputs contain multiple sentences. 
When directly decoding, NLLB-200~\cite{nllb} suffered from severe under-translation,
where the output translation only contains one sentence.
We therefore split the input by sentence boundaries and decode sentence by sentence.

\paragraph{Hyperparameters}
When decoding, we use a beam size of $4$ and length penalty of $1.0$.

\paragraph{Evaluation}
To evaluate BLEU and COMET scores, we concatenate the hypotheses and references from different attributes.
It is also the case when reporting the multi-source results in \autoref{tab:formality_src_zs}.

\subsection{Details on Finetuning}
When finetuning NLLB-200, we use a batch size of $16k$ target tokens.
For bilingual systems, we train for $30$ updates.
When training multilingually, we train for $60$ updates. 
We use a learning rate of $0.0001$ with an inverse squared root schedule and $20$ warmup steps.
Dropout is set to $0.1$.

\subsection{Details on Classifier Guidance}
\paragraph{Attribute Classifier Training}
The classifier operates on meanpooled decoder hidden states and consists of two feedforward layers with ReLU activation in between.
The first layer projects from the $1024$ Transformer hidden dimension to $256$, 
the second layer from $256$ to $C$, the number of attribute classes.
In our experiments, 
$C$ is $2$ for formality control (formal, informal) 
and $3$ for gender control (feminine, masculine, neutral)\footnote{As our test set only covers two genders, we only report scores on two genders.}.

We train the classifier on a frozen NLLB-200 600M model with an effective batch size of 32$k$ target tokens.
The learning rate is $0.002$ with an inverse square root schedule and $20$ warm-up steps.
We use the Adam~\cite{DBLP:journals/corr/KingmaB14} optimizer with betas of $(0.9, 0.98)$.
Dropout and label smoothing are set at $0.1$.
For formality control, we train the monolingual classifiers for $100$ updates and multilingual for $250$ updates.
For the gender control, we train for $25$ updates due to the small dataset and simplicity of the training data.

\paragraph{Hyperparameters}
For the classifier guidance hyperparameters, 
on the en$\rightarrow$de training data of CoCoA-MT,
we searched among step size $[0.05, 0.1, 0.5]$, 
and number of iterations $[3, 5]$.
We used $5$ iterations and $0.1$ step size for formality control, 
and $5$ iterations and $0.05$ step size for grammatical gender control. 
We do not use KL regularization and postnorm fusion as in \citet{DBLP:conf/iclr/DathathriMLHFMY20}, 
since they degraded performance in initial experiments.

\paragraph{Decoding Speed}
Decoding with our approach is slow due to the repeated gradient updates.
For instance on formality control, 
decoding on the test sets of 600 sentences takes around 30 minutes.

\section{Details on Human Evaluation} \label{appendix:human_eval_details}
We randomly sampled 16 source English sentences containing second person pronouns from the CoCoA-MT test set,
and collected 5 translations for each: from baseline NLLB-200, as well as from CG (\texttt{multi}) and FT (\texttt{multi}) for both formalities\footnote{Due to time constraints, we could not include the combination of finetuning and classifier guidance in the evaluation.}.
A native speaker rated the 80 hypotheses.

During the evaluation, we learned that there are three levels of formality in Bengali,
where: 1) the lowest formality level is only used between very close relations;
2) the next higher level is used between families or acquaintances; 
3) the highest level is used between unfamiliar persons or those between higher social distances.
We therefore asked the annotator to match each formality category to one integer point.
That is, 1, 2, and 3 correspond to very informal, informal, and formal respectively. 
We also learned that the lowest formality level is only used between very close relations and therefore rare.

While scoring, the annotator was presented with the English source sentences and their Bengali translations together in random order, 
and asked to score translation quality on a 5-point scale (1 being the worst) 
and formality scores on a 3-point scale (1 being the least formal).

\section{Comparison to Prior Works Trained on Different Data Conditions} \label{appendix:shared_tasks}

Here we compare our results 
to prior works that used \textit{more relaxed} data conditions than ours for the zero-shot tasks.
In \autoref{tab:comparison_iwslt}, first four systems are submissions to the unconstrained zero-shot track of the IWSLT 2023 formality control shared task \cite{iwslt2023_findings}.
We compare to submissions in the unconstrained track, 
as our models would fall under this track due to the use of pretrained models.
The scores of other systems are from Table 48 of \citet{iwslt2023_findings}.
We grayed out our COMET scores, 
as we are unsure whether our evaluation used the same underlying model as the organizers (we used \texttt{wmt22-comet-da}).
Overall, our model's performance is comparable to the leading systems.

\begin{table}[ht!]
\small
\centering
\setlength\tabcolsep{1pt} 
\begin{tabular}{llrrr}
\toprule
& 
\textbf{Formality} & 
\textbf{BLEU} &
\textbf{COMET} &
\textbf{M-Acc}
\\
\midrule
\textbf{en$\rightarrow$pt} \\
Ours & formal & 40.3 & {\color{gray}85.3} & 100 \\
& informal & 43.9 & {\color{gray}86.0} & 83\\
\citet{wu-etal-2023-improving} & formal & 45.4 & 77.4 & 100\\
& informal & 49.1 & 78.5 & 100\\
\citet{bahar-etal-2023-speech} & formal & 34.6 & 60.9 & 99 \\
& informal & 42.4 & 67.9 & 64 \\
\citet{lee-etal-2023-improving-formality} & formal & 31.0 & 52.5 & 100 \\
& informal & 19.9 & 24.9 & 68 \\
\citet{vakharia-etal-2023-low} & formal & 26.6 & 40.5 & 90 \\
& informal & 28.4 & 42.5 & 58 \\
\midrule
\textbf{en$\rightarrow$ru} \\
Ours & formal & 33.2 & {\color{gray}84.4} & 100 \\
& informal & 33.0 & {\color{gray}84.4} & 93\\
\citet{bahar-etal-2023-speech} & formal & 35.4 & 61.7 & 99 \\
& informal & 33.0 & 60.3 & 98 \\
\citet{wu-etal-2023-improving} & formal & 33.7 & 58.0 & 100\\
& informal & 32.4 & 55.6 & 100\\
\citet{lee-etal-2023-improving-formality} & formal & 25.8 & 44.5 & 100 \\
& informal & 26.3 & 41.8 & 100 \\
\citet{vakharia-etal-2023-low} & formal & 18.4 & -17.1 & 99 \\
& informal & 14.9 & -27.7 & 52 \\
\citet{vincent-etal-2023-mtcue} & formal & unknown & unknown &  100
\\
& informal & unknown & unknown &  99
\\

\bottomrule
\end{tabular}
\caption{\label{tab:comparison_iwslt} Comparison to prior works with different data conditions.}
\vspace{-5pt}
\end{table}

\end{document}